\documentclass{article} % For LaTeX2e
\usepackage{iclr2018_workshop,times}
\usepackage{hyperref}
\usepackage{url}
\usepackage{amssymb}

\usepackage{graphicx}
\usepackage{amsmath}
\usepackage{amsthm, mathrsfs}
\usepackage{tabu,longtable}
\usepackage{caption}
\usepackage{subfig}

\usepackage{chngcntr}

\newcolumntype{?}{!{\vrule width 1.5pt}}

\theoremstyle{plain}
\newtheorem{thm}{Theorem}
\newtheorem{lem}{Lemma}

\theoremstyle{definition}
\newtheorem{defn}{Definition}

\def\XM{{\mathscr X}}
\def\YM{{\mathscr Y}}
\def\LM{{\mathscr L}}
\def\X{{\bf X}}

\def\y{{\bf y}}
\def\Y{{\bf Y}}

\def\Enc{\mathrm{Enc}}
\def\Dec{\mathrm{Dec}}
\def\D{\mathrm{D}}
\def\Ex{\mathbb{E}}
\def\E{\mathrm{E}}

\title{DNA-GAN: Learning Disentangled Representations from Multi-Attribute Images}

\author{Taihong Xiao, Jiapeng Hong \& Jinwen Ma \\
Department of Information Science, School of Mathematical Sciences and LMAM\\
Peking University, Beijing, 100871, China\\
\texttt{\{xiaotaihong, jphong\}@pku.edu.cn} \quad \texttt{jwma@math.pku.edu.cn} \\
}

\begin{document}

\maketitle

\begin{abstract}
Disentangling factors of variation has become a very challenging problem on representation learning.
Existing algorithms suffer from many limitations,
such as unpredictable disentangling factors, poor quality of generated images from encodings,
lack of identity information, etc. In this paper, we propose a supervised learning model called DNA-GAN which
tries to disentangle different factors or attributes of images.
The latent representations of images are DNA-like, in which each individual piece (of the encoding) represents an
independent factor of the variation. By annihilating the recessive piece and swapping a certain piece of one
latent representation with that of the other one, we obtain two different representations which could be decoded into two kinds of images
with the existence of the corresponding attribute being changed.
In order to obtain realistic images and also disentangled representations, we further introduce the discriminator
for adversarial training. Experiments on Multi-PIE and CelebA datasets finally demonstrate that our proposed method is
effective for factors disentangling and even overcome certain limitations of the existing methods.
\end{abstract}

\section{Introduction}

The performance of machine learning (ML) algorithms depends on data representation because different representations
can entangle different explanatory factors of variation behind the data. Although prior knowledge can help
us design representations, the vast demand of ML algorithms in various AI domains cannot be met as feature
engineering is labor-intensive and needs domain expert knowledge. Therefore, the ML algorithms that can
automatically learn good representations of data will definitely make it easier
for people to extract useful information when building classifiers or predictors.

Among all the criteria of learning good representations discussed in \citet{DBLP:journals/pami/BengioCV13},
disentangling factors of variation is an important one that helps separate various explanatory factors.
For example, given a human-face image, we can obtain various facial characteristics about the person,
including gender, hair style, facial expression,
with/without eyeglasses and so on. However, it is quite difficult to train a single classifier which can handle
different facial characteristics or attributes entangled in a single image.
If we could obtain a disentangled representation of the face image, we can train a single classifier for multiple attributes.

In this paper, we propose a supervised method called DNA-GAN to obtain the disentangled representations of images.
The idea of DNA-GAN is motivated by the DNA double helix structure, in which different kinds of
traits are encoded in different DNA pieces, respectively.
We make a similar assumption that different visual attributes in
an image are controlled by different pieces of encodings in its latent representations.
In DNA-GAN, an encoder is used to encode an image to the attribute-relevant part and the
attribute-irrelevant part, where different pieces in the attribute-relevant part encode information of
different attributes, and the attribute-irrelevant part encodes other information.
For example, given a facial image, we are trying to obtain a latent
representation that each individual part controls different attributes, such as
hairstyles, genders, expressions and so on.
These attributes are expected to be encoded into disentangled attribute-relevant parts in
the latent representations, whereas other information such as background should be encoded
into attributes-irrelevant parts.
Through annihilating the recessive pieces and swapping certain pieces, we can obtain some novel
crossbreeds that can be decoded into new images.
With the help of the adversarial discriminator loss and the reconstruction loss, DNA-GAN can reconstruct the input
images and generate new images with new attributes.
Each attribute is disentangled from the other gradually though a process of iterative training.
Finally, we are able to obtain the disentangled representations from the latent representations.

The summary of contributions of our work is as follows:
\begin{enumerate}
	\item We propose a supervised algorithm called DNA-GAN that is able to disentangle multiple attributes as demonstrated by the experiments of
		interpolating multiple attributes on Multi-PIE~\citep{DBLP:journals/ivc/GrossMCKB10} and CelebA~\citep{DBLP:conf/iccv/LiuLWT15}
		datasets.
	\item We introduce the annihilating operation that prevents from trivial solutions: the attribute-relevant part encodes information
		of the whole image instead of a certain attribute.
	\item We employ an iterative training method to address the problem of unbalanced multi-attribute image data, which was theoretically proved
		to be more efficient than random image pairs.
\end{enumerate}

\section{Related Work}

Traditional representation learning algorithms focus on (1) probabilistic graphical models,
characterized by Restricted Boltzmann Machine (RBM)~\citep{smolensky1986information},
Autoencoder (AE) and their variants;
(2) manifold learning and geometrical approaches, such as Principal Components Analysis (PCA)~\citep{pearson1901liii},
Locally Linear Embedding (LLE)~\citep{roweis2000nonlinear}, Local Coordinate Coding (LCC)~\citep{DBLP:conf/nips/YuZG09}, etc.
However, recent researches have actively focused on developing deep probabilistic models that learn to
represent the distribution of data. Actually, \citet{DBLP:journals/corr/KingmaW13} employed an explicit model distribution
and utilized the variational inference to learn its parameters. As the generative
adversarial network (GAN)~\citep{DBLP:conf/nips/GoodfellowPMXWOCB14} was established, many implicit models have been developed.

In the semi-supervised setting, \citet{siddharth2016learning} tried to learn a disentangled representations by using an
auxiliary variable. \citet{DBLP:journals/corr/BouchacourtTN17} proposed the ML-VAE that could learn the disentangled
representations from a set of grouped observations.
In the unsupervised setting, InfoGAN~\citep{DBLP:conf/nips/ChenCDHSSA16} tries to maximize the mutual information between a small subset of latent variables and observations by introducing an auxiliary network to approximate the posterior.
However, it relies much on the a-priori choice of distributions and suffered from unstable training.
Another popular unsupervised method $\beta$-VAE~\citep{higgins2016beta}, adapted from VAE, lays
great stress on the KL distance between the approximate posterior and the prior.
However, unsupervised approaches do not anchor a specific meaning into the disentanglement.

More closely with our method, supervised methods take the advantage of labeled data and try to disentangle the factors as expected.
DC-IGN~\citep{DBLP:conf/nips/KulkarniWKT15} asks the active attribute to explain certain factor of variation
by feeding the other attributes by the average in a mini-batch.
TD-GAN~\citep{DBLP:conf/ijcai/WangWXT17} uses a tag mapping net to boost the quality of disentangled representations,
which are consistent with the representations extracted from images through the disentangling network. Besides, the
quality of generated images is improved by implementing the adversarial training strategy.
However, the identity information should be labeled so as to preserve the id information when swapping  attributes, which
renders the limitation of applying it into many other datasets without id labels.
IcGAN~\citep{DBLP:journals/corr/PerarnauWRA16} is a multi-stage training algorithm that first takes the advantage of
cGAN~\citep{DBLP:journals/corr/MirzaO14} to learn a map from latent representations and conditional information to real images,
and then learn its inverse map from images to the latent representations and conditions in a supervised manner.
The overall effect depends on each training stage, therefore it is hard to obtain satisfying images.
Unlike these models, our model requires neither explicit id information in labels nor multi-stage training.

Many works have studied the image-to-image translation between unpaired image data using GAN-based architectures, such as
\citet{DBLP:journals/corr/IsolaZZE16}, \citet{DBLP:journals/corr/TaigmanPW16}, \citet{DBLP:journals/corr/ZhuPIE17},
\citet{DBLP:journals/corr/LiuBK17} and \citet{DBLP:journals/corr/ZhouXYFHH17}. Interestingly, these models require
a form of 0/1 weak supervision that is similar to our setting. However, they are circumscribed in two image domains
which are opposite to each other with respect to a single attribute. Our model differs from them as we generalize
to the case of multi-attribute image data. Specifically, we employ the strategy of iterative training to overcome
the difficulty of training on unbalanced multi-attribute image datasets.

\section{DNA-GAN Approach}

In this section, we formally present our method.  A set $\XM$ of multi-labeled images and a set of labels $\YM$ are considered in our setting.
Let $\{(\X^1, \Y^1), \ldots, (\X^m, \Y^m)\}$ denote the whole training dataset, where $\X^i\in \XM$ is the $i$-th
image with its label $\Y^i\in \YM$. The small letter $m$ denotes the number of samples in set $\XM$ and $n$ denotes the number of attributes. The label $\Y^i = (\y_1^i, \ldots, \y_n^i)$ is a $n$-dimensional vector where each
element represents whether $\X^i$ has certain attribute or not. For example, in the case of labels with three candidates [Bangs, Eyeglasses, Smiling], the facial image $\X^i$ whose label is $\Y^i = (1,0,1)$ should
depict a smiling face with bangs and no eyeglasses.

\subsection{Model}

\begin{figure}[h]
	\centering
	\includegraphics[width=0.9\linewidth]{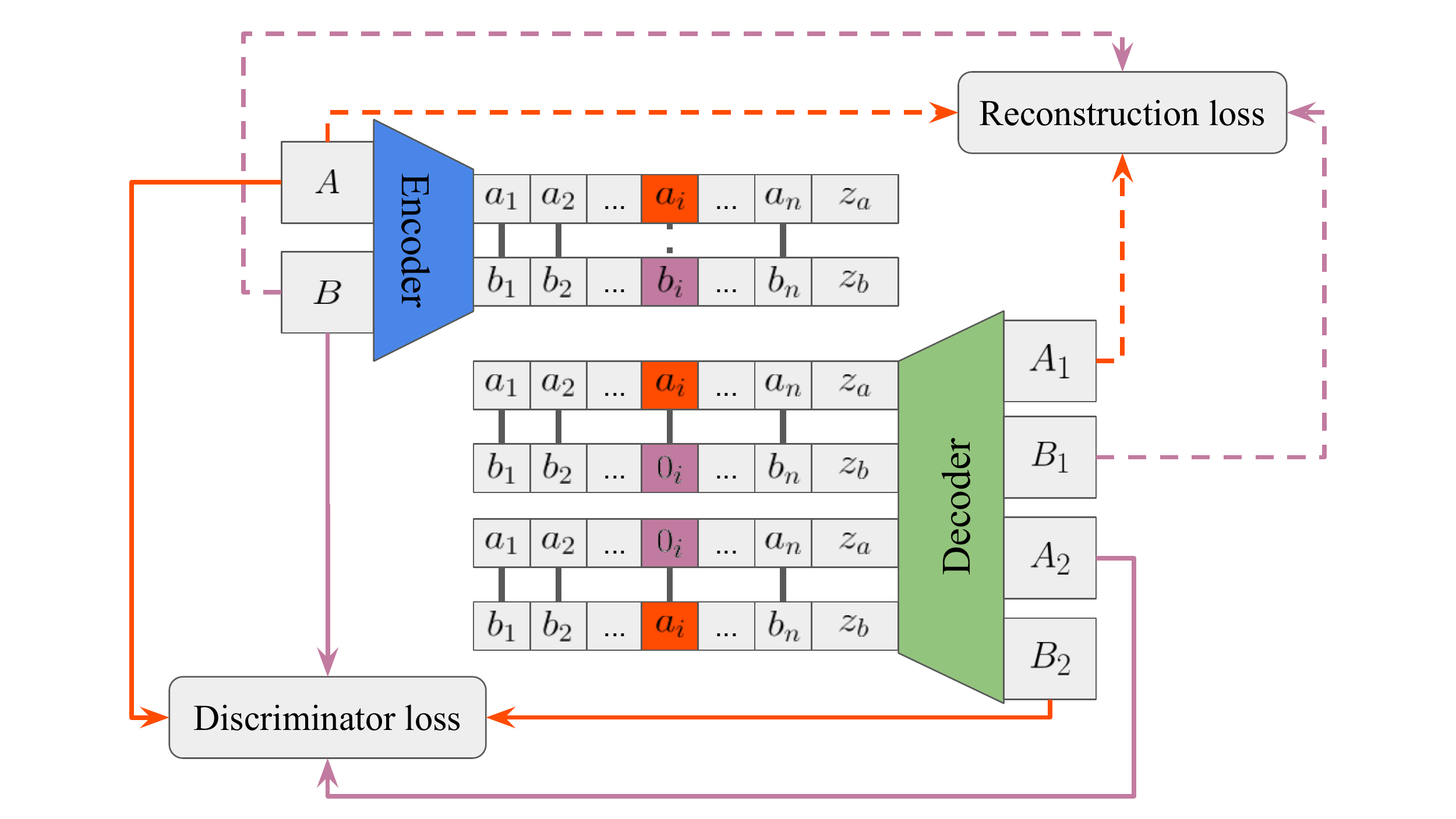}
	\caption{DNA-GAN architecture.}
	\label{DNA-GAN_framwork}
\end{figure}

As shown in Figure~\ref{DNA-GAN_framwork}, DNA-GAN is mainly composed of three parts: an encoder ($\Enc$), a decoder ($\Dec$)
and a discriminator ($\D$). The encoder maps the real-world images $A$ and $B$ into two latent disentangled representations
\begin{equation}\label{eq:encoder}
	\Enc(A) = [a_1,\ldots,a_i,\ldots,a_n,z_a], \quad
	\Enc(B) = [b_1,\ldots,b_i,\ldots,b_n,z_b]
\end{equation}
where $[a_1,\ldots,a_i,\ldots,a_n]$ is called the attribute-relevant part, and $z_a$ is called the attribute-irrelevant part.
$a_i$ is supposed to be a DNA piece that controls $\y_i$, the $i$-th attribute in the label,
and $z_a$ is for keeping other silent factors which do not appear in the attribute list as well as
image identity information.
The same thing applies for $\Enc(B)$.

We focus on one attribute each time in our framework. Let's say we are at the $i$-th attribute.
$A$ and $B$ are required to have different labels, i.e., $(\y_1^A,\ldots,1_i^A,\ldots,\y_n^A)$
and $(\y_1^B,\ldots,0_i^B,\ldots,\y_n^B)$, respectively. Under this setting, $A$ is always for the dominant pattern,
while $B$ is for the recessive pattern. We copy $\Enc(A)$ directly as the latent representation of $A_1$, and annihilate $b_i$ in the copy of $\Enc(B)$ as the latent representation of $B_1$. The annihilating operation means
replacing all elements with zeros, and plays a key role in disentangling
the attribute, which will be discussed in details in Section~\ref{sec:annihilating}. By swapping $a_i$ and $0_i$, we obtain two new latent
representations $[a_1,\ldots,0_i,\ldots,a_n,z_a]$ and $[b_1,\ldots,a_i,\ldots,b_n,z_b]$ that are supposed to be decoded into
$A_2$ and $B_2$, respectively. Via a decoder $\Dec$, we can get four newly generated images $A_1$, $B_1$, $A_2$ and $B_2$.
\begin{equation}\label{eq:decoder}
\begin{aligned}
	\Dec([a_1,\ldots,a_i,\ldots,a_n,z_a]) = A_1, &\quad \Dec([b_1,\ldots,0_i,\ldots,b_n,z_b]) = B_1 \\
	\Dec([a_1,\ldots,0_i,\ldots,a_n,z_a]) = A_2, &\quad \Dec([b_1,\ldots,a_i,\ldots,b_n,z_b]) = B_2
\end{aligned}
\end{equation}

Out of these four children, $A_1$ and $B_1$ are the direct reconstructions of $A$ and $B$, respectively,  while $A_2$ and $B_2$ are novel crossbreeds. The reconstruction losses between $A$ and $A_1$, $B$ and $B_1$, respectively, ensure the quality of directly reconstructed samples. Besides, using an adversarial discriminator $\D$ that helps make generated samples $A_2$ indistinguishable from $B$, and $B_2$
indistinguishable from $A$, we can consider attribute-related information to be encoded in $a_i$.

\subsection{Loss Functions}
\label{sec:loss}

Given two images $A$ and $B$ and their labels $\Y^A=(\y_1^A,\ldots,1_i^A,\ldots,\y_n^A)$ and
$\Y^B=(\y_1^B,\ldots,0_i^B,\ldots,\y_n^B)$ which are different at the $i$-th position,
the data flow can be summarized by \eqref{eq:encoder} and \eqref{eq:decoder}.
We force the $i$-th latent encoding of $B$ to be zero in order to prevent from trivial solutions
as we will discuss in Section~\ref{sec:annihilating}.

The encoder and decoder receive two types of losses: (1) the reconstruction loss,
\begin{align}
	L_{reconstruct} = \|A-A_1\|_1 + \|B-B_1\|_1
\end{align}
which measures the reconstruction quality after a sequence of encoding and decoding;
(2) the standard GAN loss,
\begin{align}
	L_{GAN} = -\Ex[\log(\D(A_2 | \y_i^A = 1))] - \Ex[\log(\D(B_2 | \y_i^B =0))]
\end{align}
which measures how realistic the generated images are.
The discriminator takes the generated image and the $i$-th element of its label as inputs, and outputs
a number which indicates how realistic the input image is.
The larger the number is, the more realistic the image is.
Omitting the coefficient, the loss function for the encoder and decoder is
\begin{align}
	L_G = L_{reconstruct} + L_{GAN}.
\end{align}

The discriminator $\D$ receives the standard GAN discriminator loss
\begin{align}
	L_{D_1} &= - \Ex[\log(\D(A|\y_i^A=1))] - \Ex[\log(1 - \D(B_2|\y_i^A=1))]\\
	L_{D_0} &= - \Ex[\log(\D(B|\y_i^B=0))] - \Ex[\log(1 - \D(A_2|\y_i^B=0))]\\
	L_D &= L_{D_1} + L_{D_0}
\end{align}
where $L_{D_1}$ drives $\D$ to tell $A$ from $B_2$, and  $L_{D_0}$ drives $\D$ to tell $B$ from $A_2$.

\subsection{Annihilating Operation Prevents from Trivial Solutions}
\label{sec:annihilating}

Through the experiments, we can observe that there exist trivial solutions to our model without the annihilating operation. We just take the single-attribute case as an example. Suppose that $\Enc(A) = [a, z_a]$ and $\Enc(B) = [b, z_b]$,
we can get four children without annihilating operation
\begin{equation}
	A_1 = \Dec([a,z_a]),\quad B_1 = \Dec([b,z_b]),\quad A_2 = \Dec([b,z_a]),\quad B_2 = \Dec([a,z_b]).
\end{equation}
The reconstruction loss makes it invertible from the latent encodings to the image.
The adversarial discriminator $\D$ is supposed to disentangle the attribute from other information
by telling whether $A_2$ looks as real as $B$ and $B_2$ looks as real as $A$ or not.
As is well-known, a generative adversarial network obtains the best solution when achieving the Nash equilibrium.
But without the annihilating operation, information of the whole image could be encoded intensively into the
attribute-relevant part, which just means
\begin{align}
	\Enc(A) = [a, 0], \quad \Enc(B) = [b, 0].
\end{align}
In this way, we have the following four children:
\begin{align}
	A_1 = \Dec([a, 0]),\quad B_1 = \Dec([b, 0]),\quad A_2 = \Dec([b, 0]),\quad B_2 = \Dec([a, 0]).
\end{align}
In this situation, the discriminator $\D$ cannot discriminate $A_2$ from $B$,
since they share the same latent encodings.
By the reconstruction procedure, $A_2$ and $B$ lead to the same image,
which is against our expectation that $A_2$ should depict the person from $A$ with the attribute borrowed
from $B$. The same situation appears in the case of $B_2$ and $A$ as well.

In order to get rid of these trivial solutions, we adopt the annihilating operation by replacing the recessive
pattern $b$ with a zero tensor of the same size\footnote{Use \texttt{tf.zeros\_like()} in TensorFlow implementation.}, i.e., knocking it out, and keeping the other part alive. In fact, if information of the whole image were encoded only into the attribute-relevant part, the four children in this case would be
\begin{align}
	A_1 = \Dec([a, 0]),\quad B_1 = \Dec([0, 0]),\quad A_2 = \Dec([0, 0]),\quad B_2 = \Dec([a, 0]).
\end{align}
So, the encodings of $B_1$ and $A_2$ contain no information at all and therefore the person in the corresponding images of  $B_1$ and $A_2$, who is supposed to be the same as from $B$,  cannot be reconstructed at all, which is contradictory to our aim. Hence, the attribute-irrelevant part is necessary to encode certain information of images when implementing the annihilating operation.

\subsection{Iterative Training}

To reduce the difficulty of disentangling multiple attributes, we take the strategy of iterative
training: we update our model using a pair of images with opposite labels at a certain
position each time.
Suppose that we are at the $i$-th position, the label of image $A$ is
$(\y_1^A, \ldots, 1_{i}^A, \ldots, \y_n^A)$, while the label of image $B$ is
$(\y_1^B, \ldots, 0_{i}^B, \ldots, \y_n^B)$. During each iteration, as $i$
goes through from $1$ to $n$ repeatedly, our model fed with such a pair of images can disentangle
multiple attributes one-by-one.

Compared with training with random pairs of images, iterative training is proved to be more effective.
Random pairs of images means randomly selecting pairs of images each time without label constraints.
A pair of images with different labels is called a {\it useful pair}.

We theoretically show that our iterative training mechanism is much more efficient than random image pairs
especially when the dataset is unbalanced. All proofs can be found in the \nameref{sec:appendix}.
\begin{thm}\label{th:1}
	Let $\XM = \{(\X^1,\Y^1), \ldots, (\X^m, \Y^m)\}$ denote the whole multi-attribute image dataset, where
	$\X^i$ is a multi-attribute image and its label $\Y^i=(\y_1^i,\ldots,\y_n^i)$ is an $n$-dimensional vector.
	There are totally $2^n$ kinds of labels, denoted by $\LM =\{l_1,\ldots,\l_{2^n}\}$.
	The number of images with label $l_i$ is $m_i$, and $\sum_{i=1}^{2^n} m_i = m$.
	To select all useful pairs at least once, the expected numbers of iterations needed for randomly selecting pairs and for iterative training are denoted by $\E_1$ and $\E_2$ respectively.
	Then,
	\begin{align}
		\E_1 &= m^2\left(1+\frac{1}{2}+\cdots+\frac{1}{m^2 - \sum_{i=1}^{2^n}m_i^2}\right)\\
		\E_2 &\leq 2n\cdot\max_{s=1,\ldots,n} \sum_{i\in I_s,j\in J_s}m_i m_j\left(1+\frac{1}{2}+\cdots+\frac{1}{m^2 - \sum_{k_1=1}^{2^{n-1}}(m_{i_{k_1}} + m_{j_{k_1}})^2}\right)
	\end{align}
	where $I_s$ represents the indices of labels where the $s$-th element is $1$, and $J_s$ represents the indices of labels where the $s$-th element is $0$.
\end{thm}

\begin{defn}(Balancedness)\label{defn:1}
	Define the balancedness of a dataset $\XM$ described above with respect to the $s$-th attribute
	as follows:
	\begin{equation}
		\rho_s = \frac{\sum_{i\in I_s} m_i}{\sum_{j\in J_s} m_j}
	\end{equation}
	where $I_s$ represents the indices of labels where the $s$-th element is $1$, and $J_s$ represents the
	indices of labels where the $s$-th element is $0$.
\end{defn}

\begin{thm}\label{th:2}
	We have $\E_2 \leq \E_1$, when
	\begin{equation}\label{eq:criterion}
		n\le\min_{s}\frac{(\rho_s+1)^2}{2\rho_s}.
	\end{equation}
	Specifically, $\E_2 \leq \E_1$ holds true for all $n\leq 2$.
\end{thm}

The property of the function $(\rho+1)^2/(2\rho)$ suits well with the definition of balancedness, because
it attains the same value for $\rho$ and $1/\rho$, which is invariant to different labeling methods.
Its value gets larger as the dataset becomes more unbalanced. The minimum is obtained at $\rho=1$,
which is the case of a balanced dataset.

Theorem~\ref{th:2} demonstrates that the iterative training mechanism is always more efficient than
random pairs of images when the number of attributes met the criterion \eqref{eq:criterion}.
As the dataset becomes more unbalanced, $(\rho_s+1)^2/(2\rho_s)$ goes larger, which means
\eqref{eq:criterion} can be more easily satisfied.
More importantly, iterative training helps stabilize the training process on unbalanced datasets.
For example, given a two-attribute dataset, the number of data of each kind is as follows:

\begin{table}[h]
\centering
\caption{The example of an unbalanced two-attribute dataset.}
\label{tab:1}
\begin{tabu}{c|c|c|c|c}
  \hline
  Label & $(0,0)$ & $(0,1)$ & $(1,0)$ & $(1,1)$ \\
  \tabucline{-}
  Number of data & $1$ & $1$ & $m$ & $m$ \\
  \hline
\end{tabu}

\end{table}

If $m\gg 1$ is a very large number, then it is highly likely that we will select a pair of images
whose labels are $(1,0)$ and $(1,1)$ each time by randomly selecting pairs.
We ignore the pair of images whose labels are $(1,0)$ and $(1,0)$ or $(1,1)$ and $(1,1)$,
though these two cases have equal probabilities of being chosen.
Because they are not useful pairs, thus do not participated in training.
In this case, most of the time
the model is trained with respect to the second attribute, which will cause the final learnt model
less effective to the first attribute. However, iterative training can prevent this from happening,
since we update our model evenly with respect to two attributes.

\section{Experimental Results}

In this section, we perform different kinds of experiments on two real-world datasets to validate
the effectiveness of our methods.
We use the RMSProp~\citep{DBLP:conf/icml/SutskeverMDH13} optimization method initialized by a
learning rate of 5e-5 and momentum $0$. All neural networks
are equipped with Batch Normalization~\citep{DBLP:conf/icml/IoffeS15} after convolutions or deconvolutions.
We used Leaky Relu~\citep{maas2013rectifier} as the activation function in the encoder.
Besides, we adopt strategies mentioned in Wasserstein GAN~\citep{DBLP:journals/corr/ArjovskyCB17} for stable training.
We divide all images into training images and test images according to the ratio of 9:1.
All of the following results are from test images without cherry-picking.
More details can be found at \url{https://github.com/Prinsphield/DNA-GAN}.
%The following experimental results demonstrate the advantage of our model in learning interpretable
%disentangled representations from multi-attribute data.

\subsection{Multi-PIE Database}

\begin{table}[t]
\def\arraystretch{1}
\setlength\tabcolsep{1pt}
\centering

\def\picwidth{0.7}

\begin{tabu}{ *{5}{p{0.108\textwidth}} p{0.03\textwidth} }
$A$ & $B$ & $A_2$ & $B_2$ & $A_1$ & $B_1$\\	
\end{tabu}
\includegraphics[width=\picwidth\textwidth]{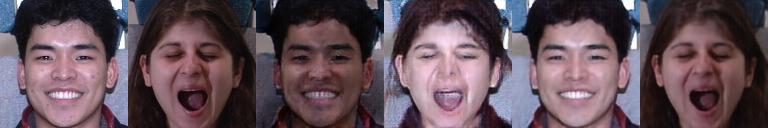}\\
\includegraphics[width=\picwidth\textwidth]{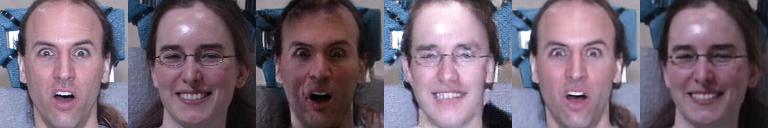}\\
\includegraphics[width=\picwidth\textwidth]{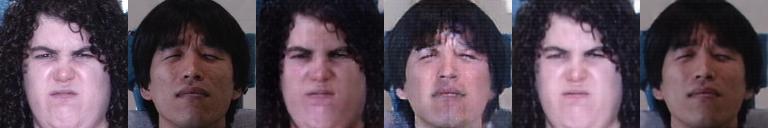}\\
\includegraphics[width=\picwidth\textwidth]{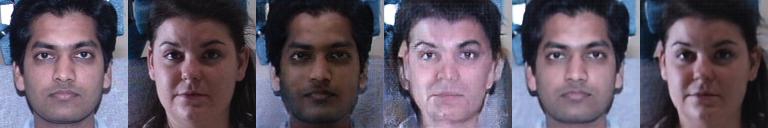}\\
\includegraphics[width=\picwidth\textwidth]{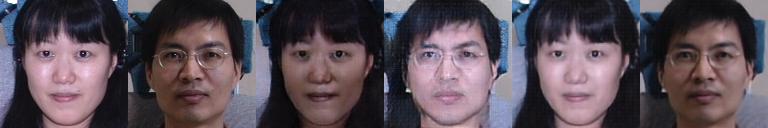}\\
\includegraphics[width=\picwidth\textwidth]{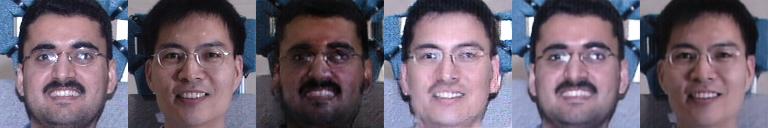}\\
\captionof{figure}{Manipulating illumination factors on the Multi-PIE dataset. From left to right,
the six images in a row are: original images $A$ with light illumination and $B$ with the dark
illumination, newly generated images $A_2$ and $B_2$ by swapping the illumination-relevant piece
in disentangled representations, and reconstructed images $A_1$ and $B_1$.}
\label{fig:multipie}
\end{table}%

The Multi-PIE~\citep{DBLP:journals/ivc/GrossMCKB10} face database contains over 750,000 images of 337 subjects
captured under 15 view points and 19 illumination conditions. We collecte all front faces images of
different illuminations and align them based on 5-point landmarks on eyes, nose and mouth.
All aligned images are resized into $128\times128$ as inputs in our experiments.
We label the light illumination face images by 1 and the dark illumination face images by 0.
As shown in Figure~\ref{fig:multipie}, the illumination on one face is successfully transferred into
the other face without modifying any other information in the images. This demonstrates that DNA-GAN
can effectively disentangle the illumination factor from other factors in the latent space.

%\subsection{3D-chairs Dataset}
%
%The 3D-chairs dataset~\citep{aubry2014seeing} contains 86366 images rendered from 1393 different chairs.
%For each chair, 62 viewpoints are taken from 31 azimuth angles and 2 elevation angles (20 and 30 degrees).
%The azimuth angle and elevations angle are two factors of variations that we are trying to disentangle.
%We collected the four images of each chair: the azimuth of 0 and 267, and the elevation angle of 20 and 30.
%The azimuth of 267 and the elevation of 30 are labeled positive, and the other two are labeled negative.
%

\subsection{CelebA Dataset}

CelebA~\citep{DBLP:conf/iccv/LiuLWT15} is a dataset composed of 202599 face images and 40 attribute binary
vectors and 5 landmark locations.
We use the aligned and cropped version and scaled all images down to $64\times 64$.
To better demonstrate the advantage of our method, we choose
TD-GAN~\citep{DBLP:conf/ijcai/WangWXT17} and
IcGAN~\citep{DBLP:journals/corr/PerarnauWRA16} for comparisons.

%In the framework of TD-GAN, the identity information should appear in the training
%label so as to preserve the id information when swapping attributes, which causes the limitation
%of applying TD-GAN into many datasets without labeling the id information.

As we mentioned before, TD-GAN requires the explicit id information in the label, thus cannot be applied to
the CelebA dataset directly.
To overcome this limitation, we use some channels to encode the id information in its latent
representations. In our experiments, the id information is preserved when swapping the attribute information
in the latent encodings.
We also compared the experimental results of IcGAN with ours in the celebA dataset. The following results
are obtained using the the official code and pre-trained celebA model
provided by the author\footnote{\url{https://github.com/Guim3/IcGAN}}.

\begin{figure}[h]
\centering

\def\picwidth{0.45}
\begin{tabu}{cc}
	
\subfloat[TD-GAN]{
\begin{minipage}[b]{\picwidth\textwidth}\label{fig:TD-GAN}
\includegraphics[width=\textwidth]{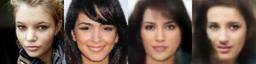}\\
\includegraphics[width=\textwidth]{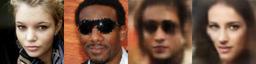}\\
\includegraphics[width=\textwidth]{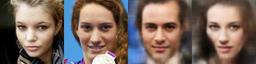}
\end{minipage}
}
&
\subfloat[IcGAN]{
\begin{minipage}[b]{\picwidth\textwidth}\label{fig:IcGAN}
\includegraphics[width=\textwidth]{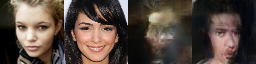}\\
\includegraphics[width=\textwidth]{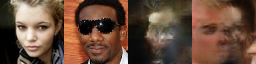}\\
\includegraphics[width=\textwidth]{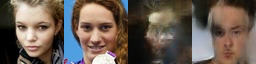}
\end{minipage}
}
\end{tabu}

\caption{The experimental results of TD-GAN and IcGAN on CelebA dataset.
Three rows indicates the swapping attributes of Bangs, Eyeglasses and Smiling.
For each model, the four images in a row are: two original images, and two newly generated images by
swapping the attributes. The third image is generated by adding the attribute to the first one,
and the fourth image is generated by removing the attribute from the second one.
}
\label{fig:comparison}
\end{figure}

As displayed in Figure~\ref{fig:TD-GAN}, modified TD-GAN encounters the problem of trivial solutions.
Without id information explicitly contained in the label, TD-GAN encodes the information of the whole image
into the attribute-related part in the latent representations. As a result, two faces are swapped directly.
Whereas in Figure~\ref{fig:IcGAN}, the quality of images generated by IcGAN are very bad, which is
probably due to the multi-stage training process of IcGAN.
Since the overall effect of the model relies much on the each stage.

\begin{figure}[t]
\centering

\def\picwidth{0.47}
\begin{tabu}{cc}

\subfloat[Bangs and Eyeglasses]{\label{fig:Bangs_Eyeglasses}\includegraphics[width=\picwidth\textwidth]{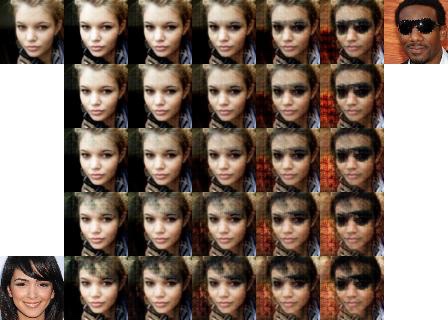}} &
\subfloat[Bangs and Smiling]{\label{fig:Bangs_Smiling}\includegraphics[width=\picwidth\textwidth]{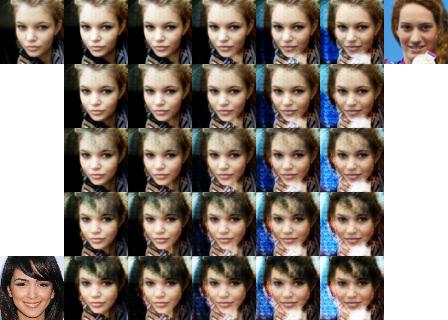}}\\
\subfloat[Eyeglasses and Smiling]{\label{fig:Eyeglasses_Smiling}\includegraphics[width=\picwidth\textwidth]{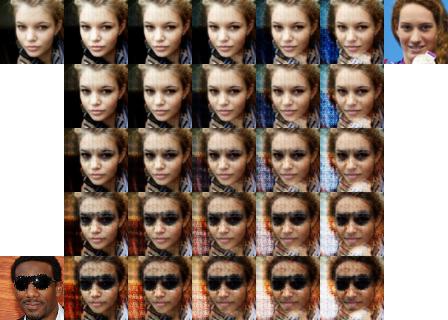}} &
\subfloat[Different Bangs]{\label{fig:Bangs}\includegraphics[width=\picwidth\textwidth]{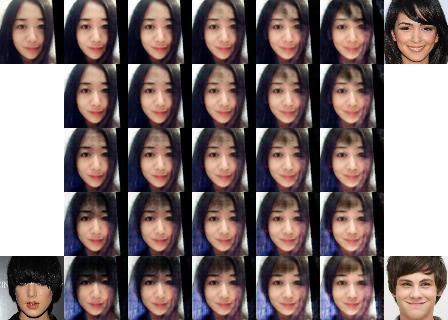}}
\end{tabu}

\caption{The interpolation results of DNA-GAN. Figure~\ref{fig:Bangs_Eyeglasses}, \ref{fig:Bangs_Smiling} and
\ref{fig:Eyeglasses_Smiling} display the disentangled
attribute subspaces spanned by any two attributes of Bangs, Eyeglasses and Smiling. Figure~\ref{fig:Bangs}
shows the attribute subspaces spanned by several Bangs feature vectors. Besides, the top-left image in
Figure~\ref{fig:Bangs} is outside the CelebA dataset. }
\label{fig:DNA-GAN}
\end{figure}

DNA-GAN is able to disentangle multiple attributes in the latent representations as shown
in Figure~\ref{fig:DNA-GAN}. Since different attributes are encoded in different DNA pieces in our latent
representations, we are able to interpolate the attribute subspaces by
linear combination of disentangled encodings. Figure~\ref{fig:Bangs_Eyeglasses}, \ref{fig:Bangs_Smiling} and \ref{fig:Eyeglasses_Smiling}
present disentangled attribute subspaces spanned by any two attributes of Bangs, Eyeglasses and
Smiling. They demonstrate that our model is effective in learning disentangled representations.
Figure~\ref{fig:Bangs} shows the hairstyle transfer process among different Bangs styles. It is worth
mentioning that the top-left image in Figure~\ref{fig:Bangs} is outside the CelebA dataset, which
further validate the generalization potential of our model on unseen data.
Please refer to Figure~\ref{fig:more_results} in the Appendix for more results.

\section{Conclusions}

We have propose a supervised model called DNA-GAN that can learn disentangled
representations from multi-attribute images. The latent representations of images are DNA-like,
consisting of attribute-relevant and attribute-irrelevant parts. By the annihilating operation and
attribute hybridization, we are able to create new latent representations which could be decoded into
novel images with designed attributes. The iterative training strategy effectively overcomes the
difficulty of training on unbalanced datasets and helps disentangle multiple attributes in
the latent space. The experimental results not only demonstrate that DNA-GAN is effective in learning disentangled
representations and image editing, but also point out its potential in interpretable deep learning,
image understanding and transfer learning. 

However, there still exist certain limitations of our model. Without strong guidance on the attribute-irrelevant parts,
some background information may be encoded into the attribute-relevant part.
As being shown in Figure~\ref{fig:DNA-GAN}, the background color gets changed when swapping attributes.
Besides, our model may fail when several attributes are highly correlated with each other.
For example, male and mustache are statistically dependent, which are hard to disentangle in the
latent representations. These are left as our future work.

\subsubsection*{Acknowledgement}

This work was supported by High-performance Computing Platform of Peking University and
the National Science Foundation of China for grant U1604153.

%\newpage
\bibliography{iclr2018_workshop}
\bibliographystyle{iclr2018_workshop}

\newpage

\section*{Appendix}
\label{sec:appendix}

\setcounter{figure}{5}

To prove Theorem~\ref{th:1}, we need the following lemma.

\begin{lem}\label{lem:1}
	A set $S=\{s_1,\ldots,s_m\}$ has $m$ different elements, from which elements are being
	selected equally likely with replacement.
	The expected number of trials needed to collect a subset $R=\{s_1,\ldots, s_n\}$ of $n(1\leq n\leq m)$
	elements is
	$$
	m\cdot\left(\frac{1}{1} + \frac{1}{2} + \cdots + \frac{1}{n}\right).
	$$
\end{lem}

\begin{proof}
	Let $T$ be the time to collect all $n$ elements in the subset $R$, and let $t_i$ be the time to collect
	the $i$-th new elements	after $i-1$ elements in $R$ have been collected. Observe that the probability
	of collecting a new element is $p_i=(n-(i-1))/m$. Therefore, $t_i$ is a geometrically distributed random variable
	with expectation $1/p_i$. By the linearity of expectations, we have:
	\begin{align*}
		\Ex(T) &= \Ex(t_1) + \Ex(t_2) + \cdots + \Ex(t_n) \\
			  &= \frac{1}{p_1} + \frac{1}{p_2} + \cdots + \frac{1}{p_n} \\
			  &= \frac{m}{n} + \frac{m}{n-1} + \cdots + \frac{m}{1} \\
			  &= m\cdot\left(\frac{1}{1} + \frac{1}{2} + \cdots + \frac{1}{n} \right).
	\end{align*}
\end{proof}

\begin{proof}{(of Theorem~\ref{th:1})}

	We first consider the case of randomly selecting pairs.
	All possible image pairs are actually in the product space $\XM \times \XM$, whose cardinality is
	$m^2$. If we take the order of two images in a pair into consideration, the number of possible pairs
	is $m^2$. Recall that the {\it useful pair} denotes a pair of image of different labels.
	Therefore, the number of all useful pairs is $\sum_{i\neq j}m_i m_j$.
	By Lemma~\ref{lem:1}, the expected number of iterations for randomly selecting pairs to select all
	useful pairs at least once is
	\begin{align}
		\E_1 &= m^2\left(1+\frac{1}{2}+\cdots+\frac{1}{\sum_{i\neq j}m_i m_j}\right)\notag\\
			 &= m^2\left(1+\frac{1}{2}+\cdots+\frac{1}{\sum_{i=1}^{2^n}(m_i\sum_{j\neq i} m_j)}\right)\notag\\
			 &= m^2\left(1+\frac{1}{2}+\cdots+\frac{1}{\sum_{i=1}^{2^n}m_i(m - m_i)}\right)\notag\\
			 &= m^2\left(1+\frac{1}{2}+\cdots+\frac{1}{m^2 - \sum_{i=1}^{2^n}m_i^2}\right).
	\end{align}
	
	Now we consider the case of iterative training. We always select a pair of images of different
	labels each time. Suppose we are selecting images with opposite labels at the $s$-th position. Let
	$I_s$ denote the indices of all labels with the $s$-th element $1$, and $J_s$ denote the indices of all
	labels with the $s$-th element $0$, where $|I_s|=|J_s|=2^{n-1}$. Then we consider the subproblem by
	neglecting the first position in data labels, the number of all possible pairs is
	$2\sum_{i\in I_s,j\in J_s}m_im_j$ (regarding of order), and the number of useful pairs is
	\begin{align}
	&\phantom{= } \sum_{k_1\neq k_2}(m_{i_{k_1}}+m_{j_{k_1}})(m_{i_{k_2}}+m_{j_{k_2}}) \notag\\
	&= \sum_{k_1=1}^{2^{n-1}}\sum_{k_2\neq k_1}(m_{i_{k_1}}+m_{j_{k_1}})(m_{i_{k_2}}+m_{j_{k_2}})\notag\\
	&= \sum_{k_1=1}^{2^{n-1}}(m_{i_{k_1}}+m_{j_{k_1}})(m - m_{i_{k_1}} - m_{j_{k_1}})\notag\\
	&= m^2 - \sum_{k_1=1}^{2^{n-1}}(m_{i_{k_1}} + m_{j_{k_1}})^2.
	\end{align}
	Therefore, the expectation to select all useful pairs at least once regardless of the $s$-th element in the label is
	\begin{align}
		\E_{\backslash s} &= 2\sum_{i\in I_s, j\in J_s}m_im_j\left(1+\frac{1}{2}+\cdots+\frac{1}{m^2 - \sum_{k_1=1}^{2^{n-1}}(m_{i_{k_1}} + m_{j_{k_1}})^2}\right)
	\end{align}
	Since we rotate the subscript $s$ from $1$ to $n$, the expected number of iterations for
	iterative training to select all useful pairs at least once is
	\begin{align}
		\E_2 &\leq n\cdot \max_{s=1,\ldots,n} \E_{\backslash s}\notag \\
		&= 2n\cdot\max_{s=1,\ldots,n} \sum_{i\in I_s,j\in J_s}m_im_j\left(1+\frac{1}{2}+\cdots+\frac{1}{m^2 - \sum_{k_1=1}^{2^{n-1}}(m_{i_{k_1}} + m_{j_{k_1}})^2}\right).
	\end{align}
	
\end{proof}

\begin{proof}{(of Theorem~\ref{th:2})}
	We firstly show that
	\begin{align}
		\sum_{k_1=1}^{2^{n-1}}(m_{i_{k_1}} + m_{j_{k_1}})^2 \ge \sum_{k_1=1}^{2^{n-1}}(m_{i_{k_1}}^2 + m_{j_{k_1}}^2) = \sum_{i=1}^{2^n}m_i^2
	\end{align}
	
	According to the result of Theorem~\ref{th:1} and the Definition~\ref{defn:1} of balancedness, we have
	\begin{align}
		\E_2 &= 2n\cdot\max_{s} \sum_{i\in I_s,j\in J_s}m_im_j\left(1+\frac{1}{2}+\cdots+\frac{1}{m^2 - \sum_{k_1=1}^{2^{n-1}}(m_{i_{k_1}} + m_{j_{k_1}})^2}\right) \notag\\
		&\le 2n\cdot\max_{s} \sum_{i\in I_s,j\in J_s}m_im_j \left(1+\frac{1}{2}+\cdots+\frac{1}{m^2 - \sum_{i=1}^{2^n}m_i^2}\right) \notag\\
		&= 2n\cdot\max_{s} \left(\sum_{i\in I_s}m_i\right) \left(\sum_{j\in J_s}m_j\right) \left(1+\frac{1}{2}+\cdots+\frac{1}{m^2 - \sum_{i=1}^{2^n}m_i^2}\right) \notag\\
		&= 2n\cdot\max_{s} \frac{\rho_s m}{\rho_s+1} \frac{m}{\rho_s+1} \left(1+\frac{1}{2}+\cdots+\frac{1}{m^2 - \sum_{i=1}^{2^n}m_i^2}\right) \notag\\
		&= \max_{s} \frac{2n\rho_s}{(\rho_s+1)^2} \cdot m^2\left(1+\frac{1}{2}+\cdots+\frac{1}{m^2 - \sum_{i=1}^{2^n}m_i^2}\right) \notag\\
		&\le \E_1.
	\end{align}
	Specifically, if $n\leq 2$,
	\begin{align}
		\frac{2n\rho_s}{(\rho_s+1)^2} \le \frac{4\rho_s}{(\rho_s+1)^2} \leq1.
	\end{align}
	The inequality holds true forever.
	
\end{proof}

%\section*{Appendix 2: More results on CelebA of DNA-GAN}

\begin{figure}[h]
\centering

\def\picwidth{0.47}
\begin{tabu}{cc}

\subfloat[Bangs and Eyeglasses]{\label{fig:Bangs_Eyeglasses01}\includegraphics[width=\picwidth\textwidth]{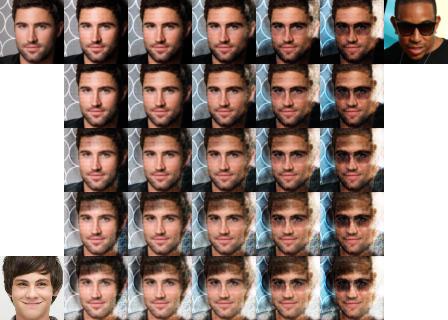}}&
\subfloat[Bangs and Eyeglasses]{\label{fig:Bangs_Eyeglasses02}\includegraphics[width=\picwidth\textwidth]{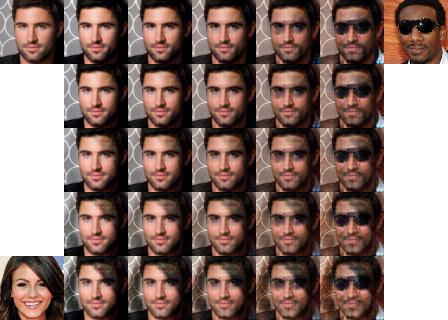}}\\

\subfloat[Bangs and Smiling]{\label{fig:Bangs_Smiling01}\includegraphics[width=\picwidth\textwidth]{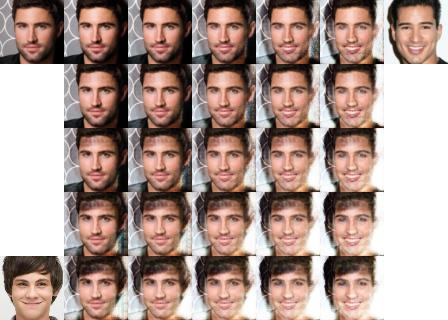}} &
\subfloat[Bangs and Smiling]{\label{fig:Bangs_Smiling02}\includegraphics[width=\picwidth\textwidth]{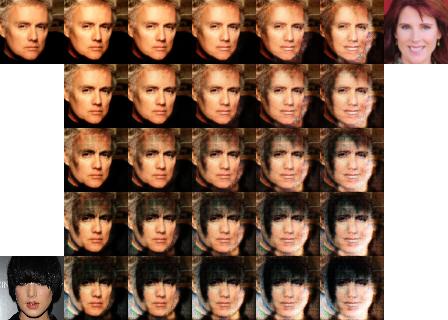}}\\

\subfloat[Male and Smiling]{\label{fig:Male_Smiling01}\includegraphics[width=\picwidth\textwidth]{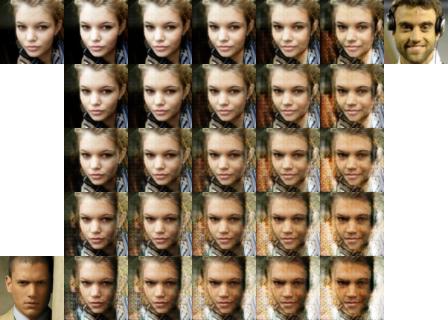}} &
\subfloat[Male and Smiling]{\label{fig:Male_Smiling02}\includegraphics[width=\picwidth\textwidth]{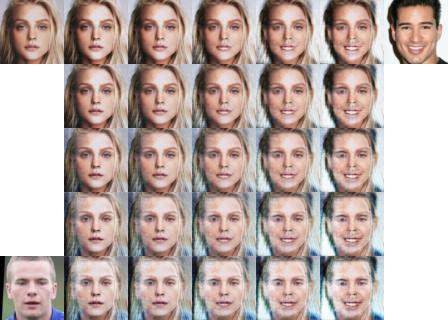}} \\

\subfloat[Male and Smiling]{\label{fig:Male_Smiling03}\includegraphics[width=\picwidth\textwidth]{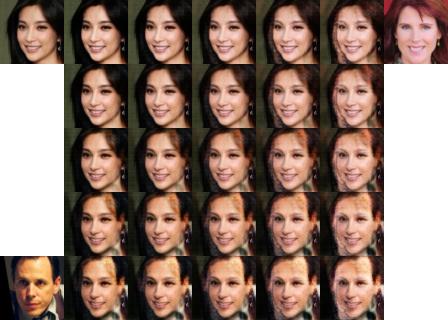}} &
\subfloat[Male and Wearing Hat]{\label{fig:Male_Wearing_Hat01}\includegraphics[width=\picwidth\textwidth]{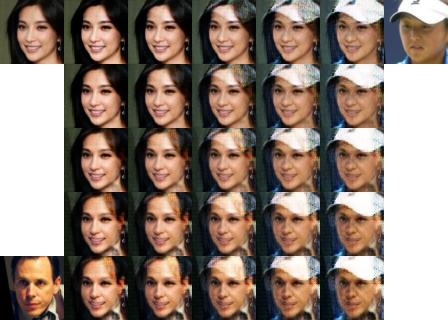}} \\
\end{tabu}
\end{figure}

\begin{figure}[h]
\ContinuedFloat
\centering

\def\picwidth{0.47}
\begin{tabu}{cc}

\subfloat[Male and Wearing Hat]{\label{fig:Male_Wearing_Hat02}\includegraphics[width=\picwidth\textwidth]{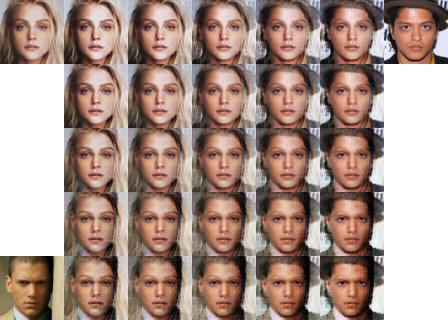}} &
\subfloat[Male and Wearing Hat]{\label{fig:Male_Wearing_Hat03}\includegraphics[width=\picwidth\textwidth]{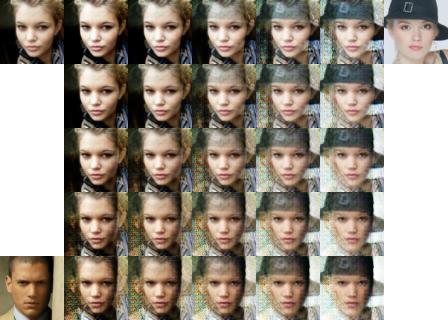}} \\

\subfloat[Smiling and Wearing Hat]{\label{fig:Smiling_Wearing_Hat01}\includegraphics[width=\picwidth\textwidth]{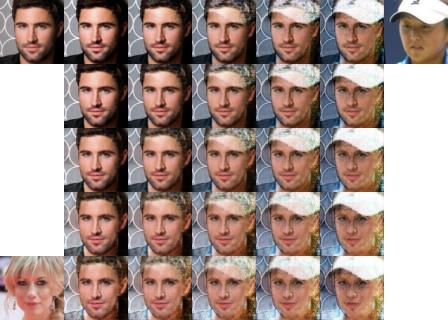}} &
\subfloat[Smiling and Wearing Hat]{\label{fig:Smiling_Wearing_Hat02}\includegraphics[width=\picwidth\textwidth]{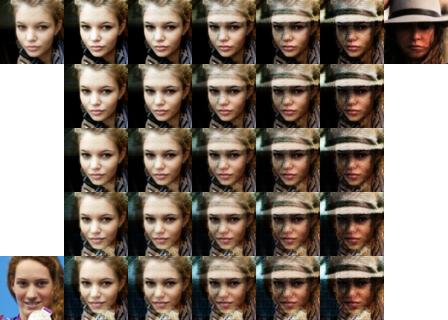}} \\

\subfloat[Smiling and Wearing Hat]{\label{fig:Smiling_Wearing_Hat03}\includegraphics[width=\picwidth\textwidth]{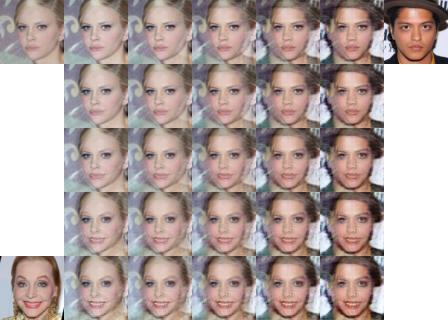}} &
\subfloat[Smiling and Wearing Hat]{\label{fig:Smiling_Wearing_Hat04}\includegraphics[width=\picwidth\textwidth]{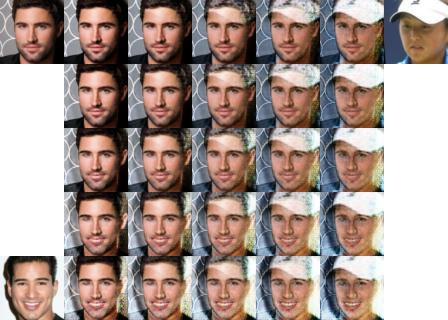}} \\

\subfloat[Wearing Hat and Mustache]{\label{fig:Wearing_Hat_Mustache01}\includegraphics[width=\picwidth\textwidth]{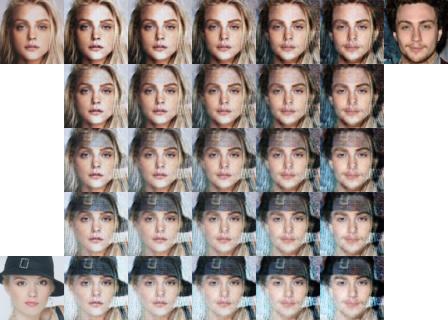}} &
\subfloat[Wearing Hat and Mustache]{\label{fig:Wearing_Hat_Mustache02}\includegraphics[width=\picwidth\textwidth]{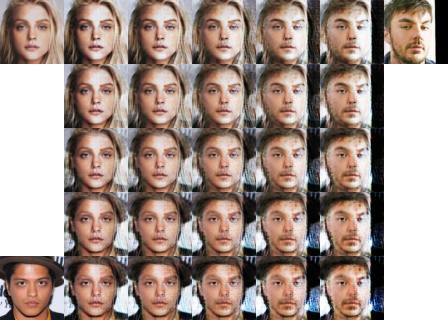}} \\

\end{tabu}

\caption{More experimental results of DNA-GAN.}
\label{fig:more_results}
\end{figure}

\end{document}